\documentclass[10pt,twocolumn,letterpaper]{article}

\usepackage[dvipsnames,table,xcdraw]{xcolor}

\usepackage{iccv}
\definecolor{iccvblue}{rgb}{0.21,0.49,0.74}

\def\confName{ICCV}
\def\confYear{2025}

\usepackage{algorithm}
\usepackage{algorithmic}
\usepackage{graphicx}
\usepackage{multirow}
\usepackage{booktabs}
\usepackage{float}
\usepackage{lipsum}
\usepackage{pifont}
\usepackage{hyperref}  
\hypersetup{
    colorlinks=true,
    linkcolor=iccvblue,
    urlcolor=iccvblue,
    citecolor=iccvblue
}

\newcommand{\cmark}{\ding{51}}%
\newcommand{\xmark}{\ding{55}}%

\def\eg{\emph{e.g.}} 
\def\ie{\emph{i.e.}}

\definecolor{DDLColor}{rgb}{1.0,0.1,0.1}

\title{DH-FaceVid-1K: A Large-Scale High-Quality Dataset for Face Video Generation}

\author{
Donglin Di$^1$\hspace{1.5mm}
He Feng$^2$\hspace{1.5mm}
Wenzhang Sun$^1$\hspace{1.5mm}
Yongjia Ma$^1$\hspace{1.5mm}
Hao Li$^1$\hspace{1.5mm}
Wei Chen$^{1}$\hspace{1.5mm}\\
Lei Fan$^3$\hspace{1.5mm}
Tonghua Su$^2$\hspace{1.5mm}
Xun Yang$^{4*}$ 
\\
\smallskip
$^1$Li Auto \hspace{4mm}
$^2$Harbin Institute of Technology \hspace{4mm}
$^3$University of New South Wales \\
$^4$University of Science and Technology of China
}

\begin{document}
\twocolumn[{%
\renewcommand\twocolumn[1][]{#1}%
\maketitle
\vspace{-1.2cm}

\begin{center}
    \centering
    \includegraphics[width=\textwidth]{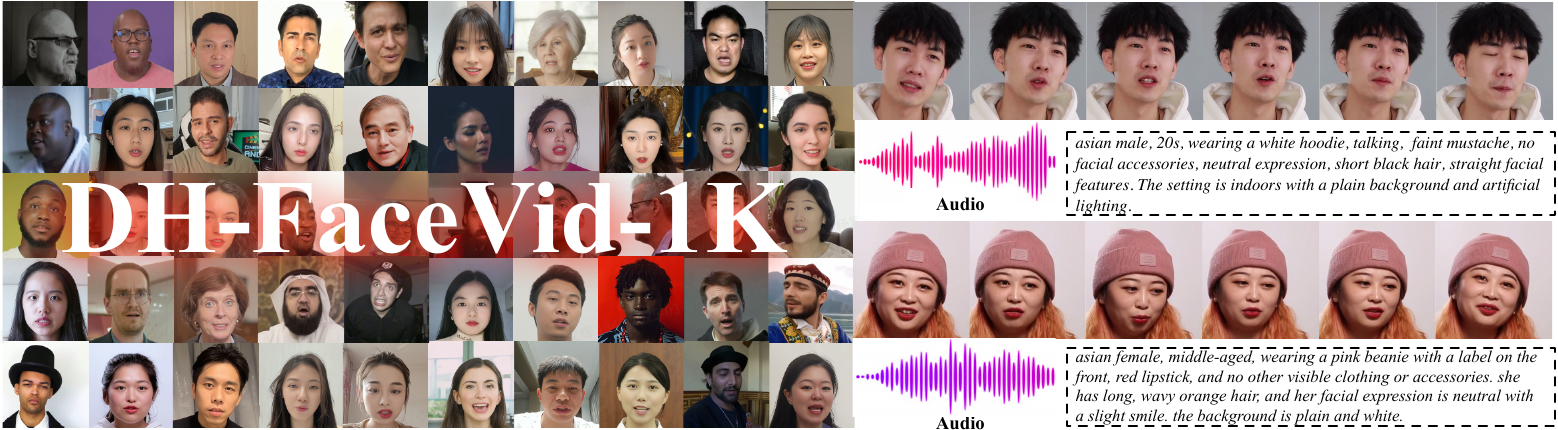}
    \vspace{-16pt}
    \captionof{figure}{
    \textbf{Overview of DH-FaceVid-1K Dataset}. It consists of 270,043 video clips along with corresponding speech audio and annotations, featuring more than 20,000 unique identities and over 1,200 hours of facial video footage captured under various environmental and lighting conditions. Notably, 80\% of the dataset represents Asian individuals, addressing the significant shortage of open-source Asian face video datasets. 
    } \label{fig:teaser}
\end{center}
}]

\begin{abstract}

\vspace{-12pt}

Human-centric generative models are becoming increasingly popular, giving rise to various innovative tools and applications, such as talking face videos conditioned on text or audio prompts. The core of these capabilities lies in powerful pre-trained foundation models, trained on large-scale, high-quality datasets. However, many advanced methods rely on in-house data subject to various constraints, and other current studies fail to generate high-resolution face videos, which is mainly attributed to the significant lack of large-scale, high-quality face video datasets. In this paper, we introduce a human face video dataset, \textbf{DH-FaceVid-1K}. Our collection spans 1,200 hours in total, encompassing 270,043 video clips from over 20,000 individuals. Each sample includes corresponding speech audio, facial keypoints, and text annotations. Compared to other publicly available datasets, ours distinguishes itself through its multi-ethnic coverage and high-quality, comprehensive individual attributes. We establish multiple face video generation models supporting tasks such as text-to-video and image-to-video generation. In addition, we develop comprehensive benchmarks to validate the scaling law when using different proportions of proposed dataset. Our primary aim is to contribute a face video dataset, particularly addressing the underrepresentation of Asian faces in existing curated datasets and thereby enriching the global spectrum of face-centric data and mitigating demographic biases.
\textbf{Project Page:} \url{https://luna-ai-lab.github.io/DH-FaceVid-1K/}

\end{abstract}

\vspace{-15pt}

\section{Introduction}
\vspace{-4pt}

Generating face videos \cite{xu2024vasa,cui2024hallo2} has become one of the most popular video generation tasks. There are various downstream tasks for creating talking face videos under different input conditions, such as using text prompts \cite{li2021write,zhang2022text2video,wang2023text,jang2024faces}, reference images \cite{hong2023implicit,chu2024gpavatar,ye2024real3dportrait}, driving audio \cite{liu2023moda,gan2023efficient,tan2024style2talker,tan2024say,drobyshev2024emoportraits,cui2024hallo3},
or facial landmarks \cite{feng2024one,liang2024emotional}.
Current research on face video generation primarily follows two major branches: GAN-based methods \cite{goodfellow2014generative,guo2024liveportrait,siarohin2021motion,siarohin2019first,wang2021one,zhao2022thin,hong2022depth} and diffusion-based methods \cite{ho2020denoising,song2021denoising,zeng2023face,yang2024megactor,xie2024x,wei2024aniportrait}.
Recently, diffusion-based methods have gained increasing popularity, offering higher quality, richer diversity, and more natural head movements \cite{dhariwal2021diffusion}.


Current diffusion-based methods \cite{shen2023difftalk,stypulkowski2024diffused,wei2024aniportrait,ma2024follow} primarily focus on adopting pre-trained universal generation backbone models, \eg, Stable Diffusion \cite{rombach2022high}, Diffusion Transformer \cite{peebles2023scalable,hong2022cogvideo,yang2024cogvideox}, and Stable Video Diffusion \cite{blattmann2023stable}, leveraging their generative ability and further optimizing the conditions to generate talking face videos, which is similar to Text-to-Image \cite{ramesh2021zero} and Text-to-3D \cite{ding2024text,di2025hyper} generation.

Drawing lessons from the remarkable success of Large Language Models (LLMs) \cite{touvron2023llama,hu2024psycollm,pan2025precise} and video generation/analysis approaches \cite{wang2023all,jin2024video,ma2025adams,yang2021deconfounded,yang2022video,yang2024robust,Zhou_2025_CVPR}, high-quality, domain-specific, publicly available large-scale datasets \cite{fan2022grainspace,fan2023annotated,fan2025manta} play a crucial role in enhancing model's generalization, robustness, and generation capabilities.
However, in the field of human face video generation, current methods often overlook the potential performance improvement brought by high-quality datasets and corresponding pre-training strategies.
Meanwhile, widely-adopted public face video datasets (\eg, HDTF \cite{zhang2021flow}, CelebV-HQ \cite{zhu2022celebv}, CelebV-Text \cite{yu2023celebv}) suffer from three following critical limitations:
\begin{itemize}
    \item \textbf{Limited total duration}. Most datasets contain fewer than 300 hours of footage, which is insufficient for pre-training backbone models in face video generation;
    \item \textbf{Quality-quantity trade-off}. Existing datasets struggle to balance data scale and quality. For instance, VoxCeleb2  \cite{chung2018voxceleb2} offers an extended duration but only provides a relatively low-resolution of $224 \times 224$ pixels, while TalkingHead-1KH \cite{wang2021one} has similar constraints;
    \item \textbf{Insufficient diversity}. The underrepresentation of certain ethnic groups, particularly Asians, limits the model's capability in generating diverse face videos.
\end{itemize}

In this paper, to mitigate these challenges, we collected diverse face videos sourced from a crowdsourcing platform, producing nearly 1,000 hours with a strict annotation procedure, and we further included 200 hours of curated public dataset content to enhance ethnic diversity for supplementing our dataset. This effort results in a comprehensive, high-fidelity, multimodal human facial video dataset, named \textbf{DH-FaceVid-1K}, as depicted in Figure \ref{fig:teaser}. This dataset comprises over 1,200 hours of video,
encompassing 270,043 video samples from over 20,000 individuals, where 65,353 samples are collected from trimmed public datasets \cite{yu2023celebv} and 204,690 samples from our own collected and filtered data. All videos are annotated with the corresponding speech audio, facial keypoints, and detailed facial annotations describing the speaker (\eg, ethnicity, gender, emotion, environment, lighting condition, and hair color). Each video has a resolution exceeding $512 \times 512$, with 46.5\% of these videos having a resolution of or greater than $1080 \times 1080$. Compared to other datasets, DH-FaceVid-1K excels in balancing scale, quality, and diversity, featuring a wide range of ethnicities and emotions. Notably, Asian faces constitute approximately 80\% of the dataset, addressing the prior underrepresentation in existing resources.

We conducted extensive experiments, including Text-to-Video (T2V) and Image-to-Video (I2V), to compare the performance of models trained on our dataset with those trained on public datasets under the same conditions, thereby validating the superiority of the proposed dataset. Furthermore, to explore the impact of dataset scale on the performance of face video generation backbones and the differences between these backbones, we addressed two key research questions (\textit{RQ}) that require empirical investigation: \textit{RQ1: Data Scale Requirement.} What is the cost-effective required scale of high-quality data, and what is the corresponding number of model parameters needed to train a well-performing backbone model for a specific domain? \textit{RQ2: Differences in Backbones.} How do various widely adopted video generation models compare in terms of performance and computational cost under different conditions for talking face video generation? To explore them, we fine-tune a state-of-the-art video generation backbone \cite{yang2024cogvideox} with varying parameter sizes (2B \& 5B) using subsets of the proposed dataset at different data scales. This analysis is conducted under both T2V and I2V settings, providing a comprehensive assessment of data scale and model performance.
The contributions can be summarized as follows:
\begin{itemize}
\item We introduce DH-FaceVid-1K, a large-scale talking face video dataset with over 1,200 hours of content, exceeding all currently available public datasets and serving as a comprehensive research resource.
\item Our dataset features multi-ethnic diversity and high-resolution video quality, effectively mitigating the underrepresentation of Asian faces prevalent in existing datasets and ensuring the robust performance of face video generation backbone across different ethnicities.
\item Building upon our dataset, extensive experiments are conducted to demonstrate that these pre-trained models enhance performance on related tasks, establishing a series of new benchmarks for future research.
\end{itemize}

\begin{figure}[!t]
    \centering
    \includegraphics[width=\linewidth]{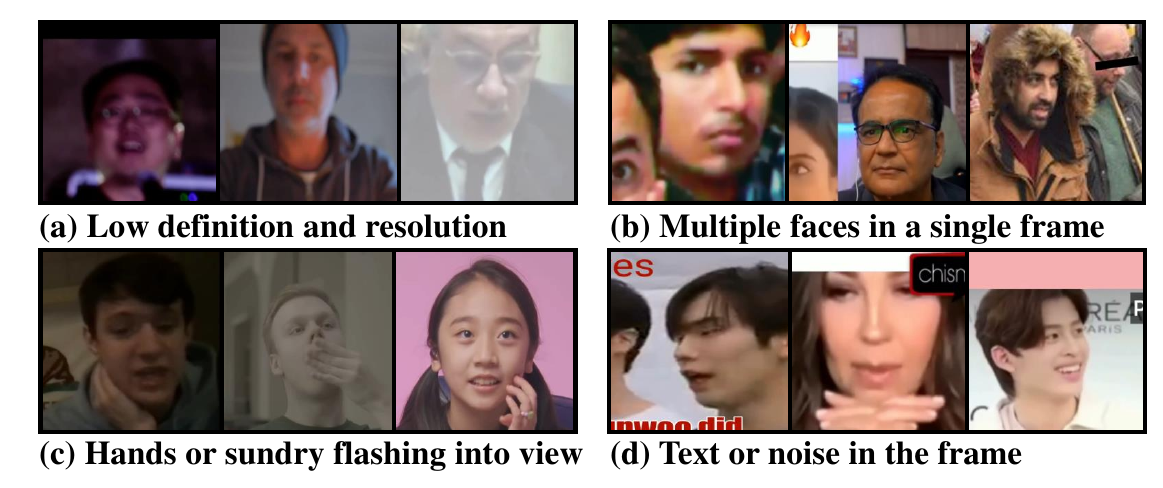}
    \vspace{-16pt}
    \caption{We identify several common issues in existing public human face video datasets that significantly contribute to the poor quality of videos generated by the corresponding trained models. These issues have been largely neglected in previous works.}
    \vspace{-15pt}
    \label{fig:dataset_prob}
\end{figure}

\begin{figure*}[t]
    \centering
    \includegraphics[width=\textwidth]{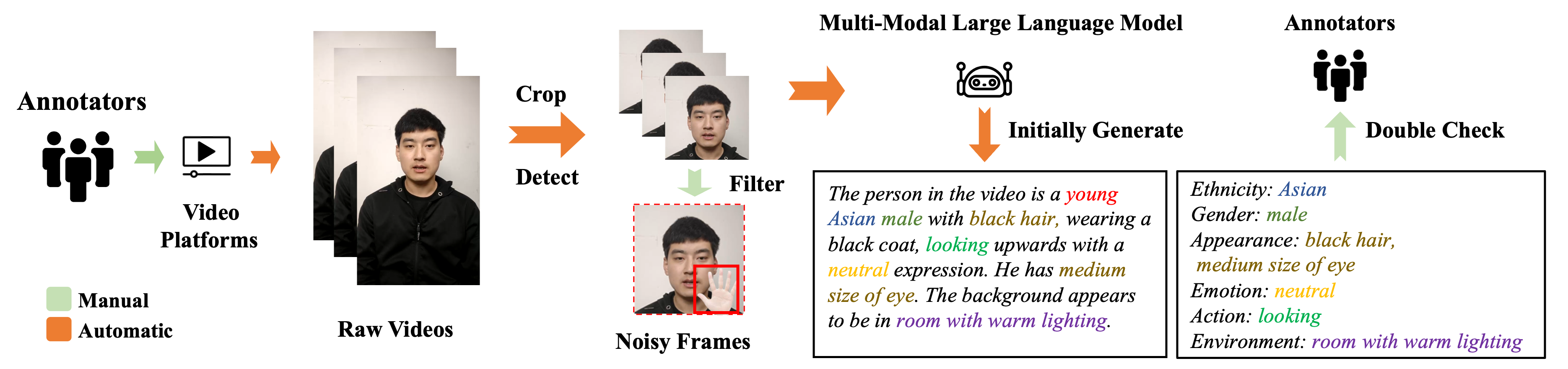}
    \caption{Illustration of the processing pipeline for collecting DH-FaceVid-1K dataset, which includes the following main steps: collecting raw videos, detecting and cropping to the face region, filtering out noisy clips, and generating video descriptions.}
    \label{fig:process}

\end{figure*}

\section{DH-FaceVid-1K Dataset}

In this section, we first provide detailed information on the collection and pre-processing methods employed.
Then, we present the statistical properties of our collected dataset and compare it with those of other public datasets.

\subsection{Collection, Processing, and Annotations}

\textbf{Collection.} We aim to build a large-scale, high-quality face video dataset. To ensure that the collected videos contain clear facial features without other human body parts or unwanted noise, we primarily gathered content from interview programs and vlogs, as these videos are typically single-person, filmed in professional environments with good lighting, high-quality equipment, and stable shooting techniques.
All raw videos were sourced from crowdsourcing platforms with proper permissions, and all identifiable information was removed. We selected raw videos with resolutions exceeding 1080p, totaling about 2,000 hours.
The raw videos were initially cropped to include the full face and the upper shoulder area. We then utilized FaceXFormer \cite{narayan2024facexformer} to automatically filter out face videos of individuals younger than 22 years old. Additionally, we used OpenCV to ensure that the face region is at least $256 \times 256$ pixels.

\noindent \textbf{Processing and Annotations.} We first drew inspiration from existing public datasets, such as CelebV-Text, and identified several common drawbacks that impose significant constraints on data quality. For example, these limitations include low definition and resolution, multiple faces in a single frame, hands or other objects momentarily obstructing the view, and text or noise within the frame, as illustrated in Figure \ref{fig:dataset_prob}.
To mitigate these issues while ensuring high dataset quality, we not only developed a data processing pipeline, as depicted in Figure \ref{fig:process}, but also employed over 100 annotators to manually annotate labels, double-check, and cross-check the annotations over six months.

We initially filtered out videos containing noise information. For subtitle detection, five frames were randomly sampled from each video, including the first and last frames. These frames were converted to grayscale and then binarized. Subsequently, contours in the images were detected to identify potential subtitle regions. Optical Character Recognition (OCR) \cite{shi2016end, paddleocr2020} was applied to these regions, and if the extracted text exceeded 10 characters in length, the presence of subtitles was confirmed.
For black border noise, we detected continuous black pixels along the top, bottom, left, and right of the video, treating sections exceeding 20 pixels as noise.
For multiple faces, we used OpenCV to detect the number of faces and discard all videos with more than one face.
To address random hand appearances in videos, we manually filtered out videos containing hands, text subtitles, or other extraneous parts.
After obtaining filtered videos, some videos remained blurry. To enhance the quality of these videos (accounting for less than 4\% of total dataset), we applied the CodeFormer \cite{zhou2022codeformer} method.

\begin{figure*}[t]
    \centering
    \includegraphics[width=.98\textwidth]{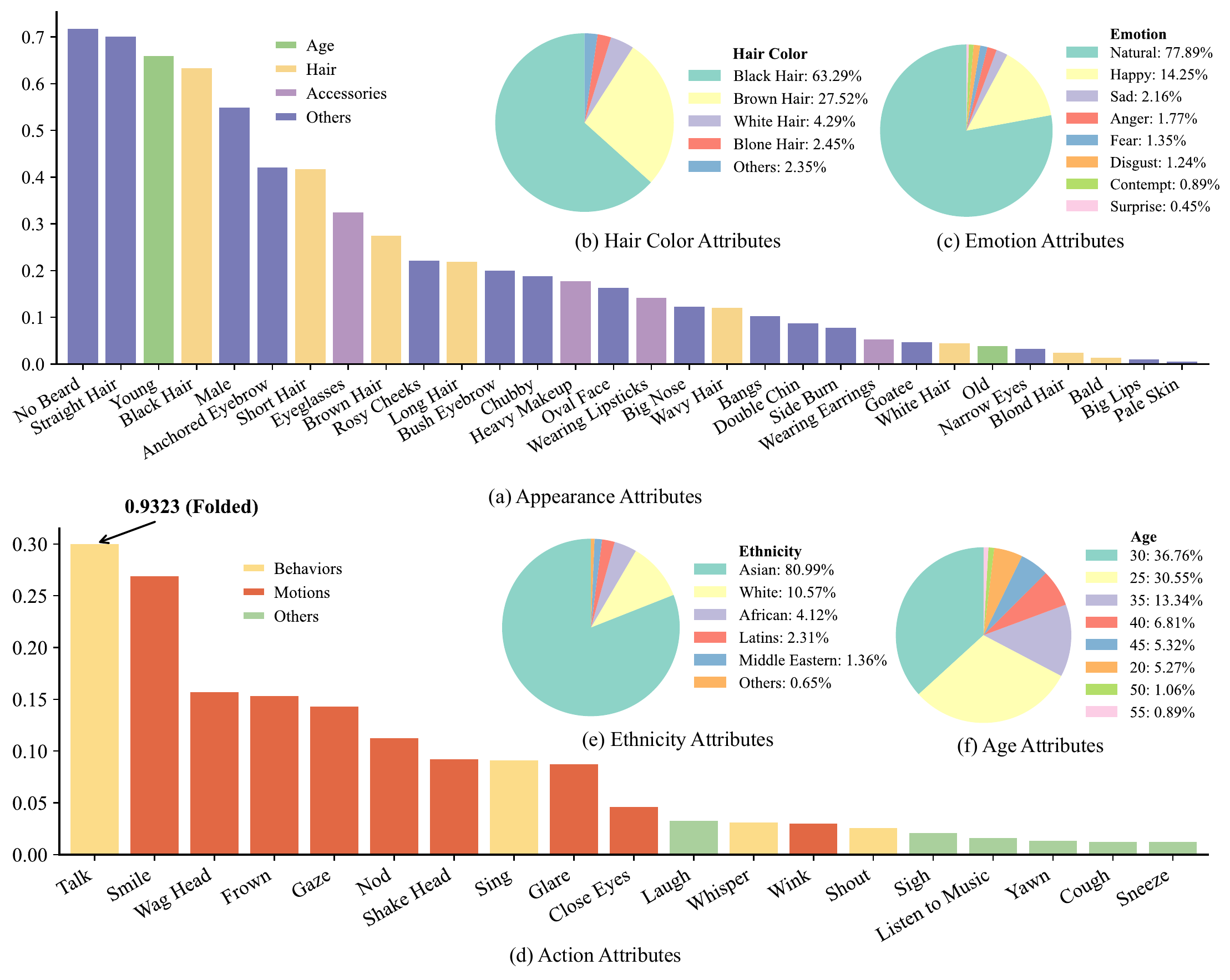}
    \caption{Distributions of general appearances in \textbf{DH-FaceVid-1K}, including hair colors, emotions, actions, ethnicities, and ages.}
    \label{fig:figure4}
\end{figure*}



We used DWPose \cite{yang2023effective} to extract facial keypoints from all videos, and further proceeded to generate detailed video annotations. Specifically, we utilized a pre-trained video captioning model, PLLaVA \cite{xu2024pllava}, to automatically generate initial annotations.
Unlike CelebV-Text \cite{yu2023celebv}, which employs description templates for semi-automatic annotation generation, we leveraged prompt-based techniques similar to those used in LLM interactions to enhance effectiveness. The captioning model was tasked with generating single-sentence descriptions of the characters in the videos, detailing their ethnicity, gender, age, appearance attributes (\eg, hairstyle and hair color), emotion, actions (predominantly talking), background environment, and lighting conditions. To capture the dynamic actions of characters in the videos (\ie, head movements), we randomly sampled 3 non-consecutive frames while generating captions to balance text accuracy and processing efficiency. Finally, we manual double-checked all annotations to ensure consistency, standardization, and high quality.
To be specific, our manually double-checking process can be divided into two stages, each involving separate annotator groups.
In the first stage, samples were checked against ISO 2859 standards to ensure all annotations align with the video content, including both static attributes (\eg, emotion and facial features) and dynamic aspects (\eg, lighting, head poses and movements).
Multiple annotators cross-verified each sample and flagged any discrepancies.
In the second stage, the flagged samples underwent further review to confirm their usability.
Throughout this process, an online platform with form-based checks ensured consistency and quality control.

These rigorous processes resulted in a high-quality self-collected data spanning over 1,000 hours of video, with 95\% of the clips depicting Asian faces. 
To further enhance data diversity, We applied the same pipeline to clean CelebV-HQ, CelebV-Text, and TalkingHead-1KH, resulting in 9,841, 32,595, and 22,917 samples respectively, totaling 200 hours. In summary, our \textbf{DH-FaceVid-1K} dataset contains nearly 1200 hours by combining these processed videos.

\textbf{Data Sources and Agreement}. 
We collected raw data from crowdsourcing platforms to ensure both the scale and diversity of our dataset, while simultaneously mitigating potential data ethics concerns. We comply with stringent legal and usage policies, applying thorough screening processes and anonymization measures to safeguard privacy. These practices help prevent misuse, including unauthorized face identification and deepfake creation.

\begin{table}[!t]
\centering
\caption{Comparison of our \textbf{DH-FaceVid-1K} with five related public datasets. A black tick denotes partial audio presence.}

\resizebox{\columnwidth}{!}{%
\begin{tabular}{l|c>{\centering\arraybackslash}m{2.5cm}|c|cc|c}
\toprule
Datasets & \#. Samples & Resolution & Duration & \multicolumn{2}{c|}{Text} & Audio \\
\midrule
VoxCeleb2 & 1,092,008 & 224$\times$224 & 2400h & \multicolumn{2}{c|}{\xmark} & \textcolor{red}{\cmark} \\
HDTF & 368 & 512$\times$512 $+$ & 15.8h & \multicolumn{2}{c|}{\xmark} & \textcolor{red}{\cmark} \\
TalkingHead-1KH & 80,000 & 512$\times$512 $-$ & 160h & \multicolumn{2}{c|}{\xmark} & \cmark \\
CelebV-HQ & 35,666 & 512$\times$512 $+$ & 68h & \multicolumn{2}{c|}{\textcolor{red}{\cmark}} & \cmark \\
CelebV-Text & 70,000 & 512$\times$512 $+$ & 279h & \multicolumn{2}{c|}{\textcolor{red}{\cmark}} & \cmark \\
\midrule
\textbf{DH-FaceVid-1K} & \textbf{270,043} & \textbf{512$\times$512 $+$} & \textbf{1200h} & \multicolumn{2}{c|}{\textcolor{red}{\cmark}} & \textcolor{red}{\cmark} \\
\bottomrule
\end{tabular}%
}
\vspace{-5pt}
\label{tab:meta}
\vspace{-10pt}
\end{table}

\subsection{Dataset Statistics}
\textbf{Data Scale.} The DH-FaceVid-1K dataset contains 270,043 video clips, comprising 65,353 clips sourced from trimmed public datasets and 204,690 newly collected clips. As illustrated in Table \ref{tab:meta}, while VoxCeleb2 offers a substantial total duration, its data resolution remains relatively low, a limitation also observed in TalkingHead-1KH. In contrast, HDTF, CelebV-HQ, and CelebV-Text provide higher-quality data, yet their total durations are somewhat limited.
CC \cite{hazirbas2021towards} and CCv2 \cite{porgali2023casual} provide word-level annotations for spoken utterances. However, CC's audio content consists of responses to random questions from a pre-approved list, while CCv2's audio content includes either a sample paragraph from ``The Idiot" by Fyodor Dostoevsky or answers to one of five predetermined questions. The relatively monotonous audio content in both datasets hinders the performance and generalizability of audio-driven talking face generation models.
Our dataset successfully balances data quality and quantity, making it a comprehensive resource for face video research.

\noindent \textbf{Data Diversity.}
We define data diversity to encompass a rich array of ethnicity and facial appearance, varying age groups and emotions, actions beyond talking, as well as different head poses, environments, and lighting conditions.

DH-FaceVid-1K specifically addresses the underrepresentation of Asian faces in video datasets, comprising over 80\% Asian faces complemented by 11\% White, 4\% African, and 5\% other ethnicities (see Figure \ref{fig:figure4}e). The dataset maintains gender balance (55\% male, 45\% female) and covers four age groups: minors (15\%), adults (65\%), middle-aged (18\%), and elderly (2\%). Appearance attributes exhibit a long-tail distribution of 30 distinct features, with Asian-associated traits such as black straight hair being the most prevalent (see Figure \ref{fig:figure4}a-b). Eleven attributes occur in over 20\% of the dataset (\eg, No Beard, Young), while eight fall between 10-20\% prevalence (\eg, Bush Eyebrow, Wavy Hair).
For emotion distributions, as shown in Figure \ref{fig:figure4}c, Neutral (65\%) is the most common emotion, largely due to the conversational nature of the content, followed by Happy (20\%) and Angry (8\%). Talking (70\%) is the predominant action, supplemented by conversational gestures such as Smiling (15\%) and Head Movements (10\%), along with specialized actions like Singing (3\%), as shown in Figure \ref{fig:figure4}d. 

For the head pose distributions, we followed pipeline from EFHQ \cite{dao2024efhq}. Specifically, we uniformly sampled 5 frames from each video to estimate the head poses and categorized them into the following 6 classes: frontal (74\%), profile\_right (11\%), profile\_left (8\%), profile\_up (3\%), profile\_down (3\%), and extreme (1\%).
Since there may be head pose variations between frames, we assigned the dominant pose category as the video-level pose annotation.

\begin{figure}[t]
    \centering
    \includegraphics[width=.98\columnwidth]{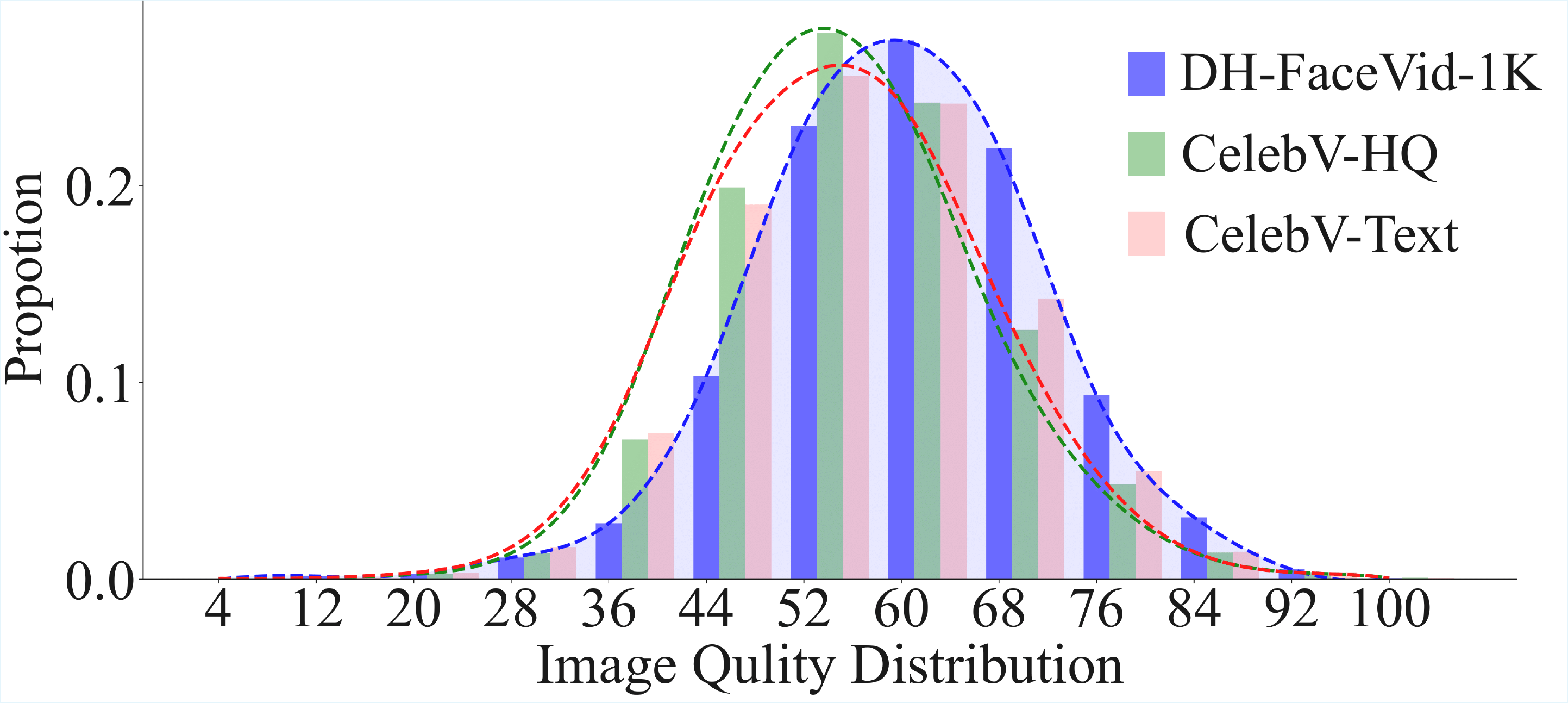}
    \caption{Comparison of image quality between our dataset and the highest-quality public datasets (CelebV-HQ and CelebV-Text).}
    \label{fig:quality}
\end{figure}

\noindent \textbf{Data Quality.} We followed the evaluation approach used in CelebV-Text \cite{zhu2022celebv} to measure video quality, employing BRISQUE \cite{mittal2012no} to assess the quality of each video frame and then averaging the results. As shown in Figure \ref{fig:quality}, our DH-FaceVid-1K achieves a higher BRISQUE score than CelebV-HQ and CelebV-Text, highlighting its superior image quality. This improvement stems largely from our comprehensive data processing pipeline and extensive manual filtering, which together ensure a high-quality dataset.

\begin{figure}[t]
    \centering
    \includegraphics[width=.98\columnwidth]{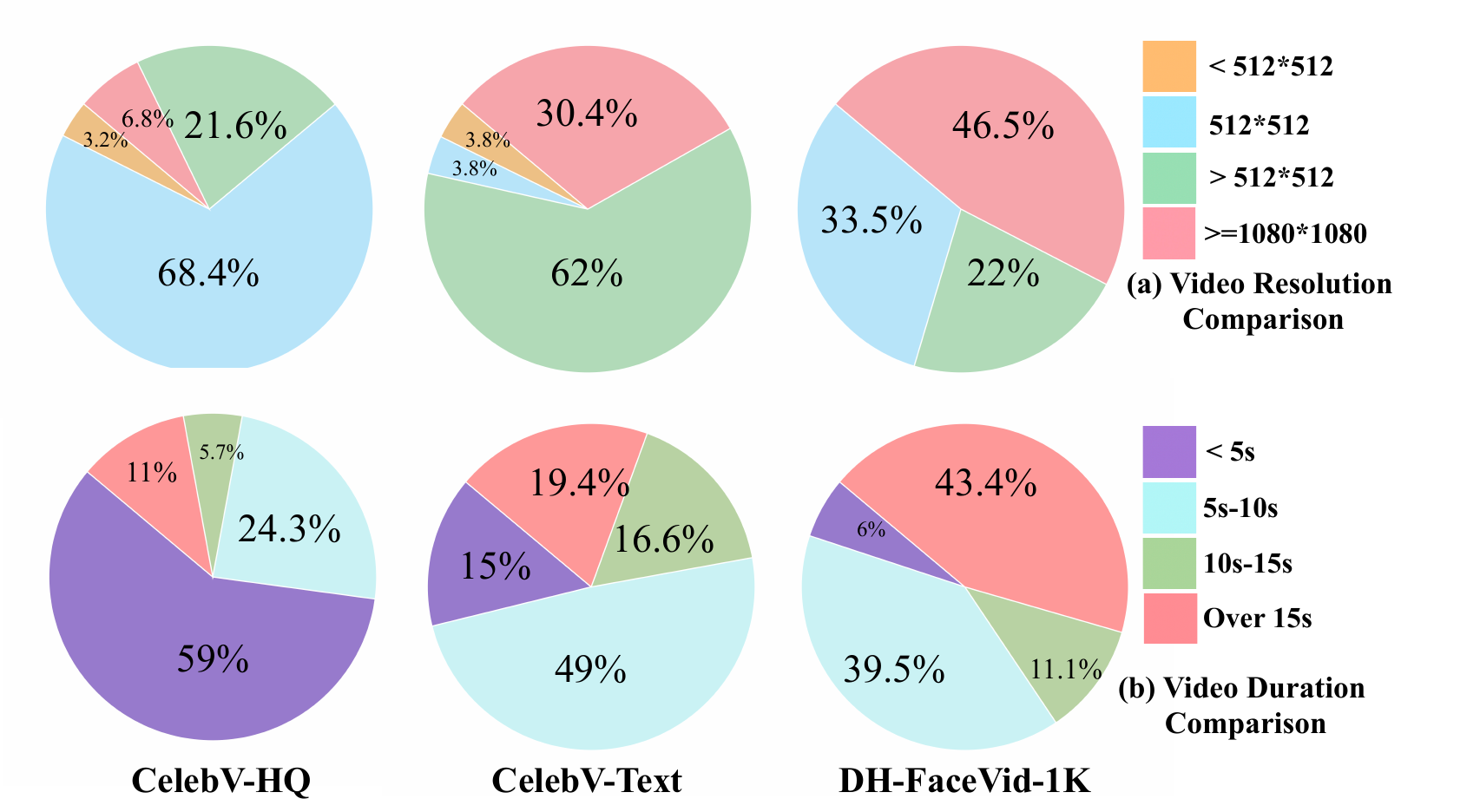}
    \caption{Video resolution and duration comparison among CelebV-HQ, CelebV-Text, and DH-FaceVid-1K.}
    \label{fig:piechart}
\end{figure}

We analyzed the duration and resolution distribution of our dataset, as shown in Figure \ref{fig:piechart}, DH-FaceVid-1K offers significantly higher-resolution and longer videos than CelebV-HQ and CelebV-Text: 46.5\% of its clips are in 1080p (vs. 6.8\% and 30.7\%, respectively), and 43.4\% exceed 15 seconds in duration (vs. 11.0\% and 19.4\%). These longer videos benefit face video generation training by improving visual quality and temporal consistency.

\subsection{Audio Filtering}
To further enhance the dataset's quality and maximize its potential, we applied additional audio filtering to support audio-driven talking head generation. Ensuring accurate alignment between the video’s audio and lip movements is paramount. Notably, the SyncNet \cite{chung2017out} pre-trained on LRW \cite{chung2017lip} showed significant scoring deviations for non-English speech. Consequently, we adopted \cite{li2024latentsync} to retrain a new SyncNet, which we then used to generate SyncNet Scores for each video. Samples exhibiting poor lip-audio synchronization were filtered out to ensure overall dataset integrity.

\section{Data Safety} 
To prevent potential misuse of DH-FaceVid-1K such as in face identification, deepfake, or generating videos that perpetuate stereotypes or biases, we have instituted strict data governance protocols.
\begin{table*}[!htbp]
\LARGE
\setlength\tabcolsep{4pt}
    \centering
    \caption{Quantitative Results on Text-to-Video Generation. We evaluate the performance of five state-of-the-art open-source models for this task, maintaining consistent settings to facilitate fair comparison across different datasets.}
    \vspace{-8pt}
    \label{tab:quantitative-t2v}
    \resizebox{\textwidth}{!}{ 
    \begin{tabular}{l|ccc|ccc|ccc|ccc|ccc}
        \toprule
        \multicolumn{1}{l|}{Methods} & \multicolumn{3}{c|}{AnimateDiff \cite{guo2023animatediff}} & \multicolumn{3}{c|}{Latte \cite{ma2024latte}} & \multicolumn{3}{c|}{OpenSora \cite{opensora}} & \multicolumn{3}{c|}{EasyAnimate \cite{xu2024easyanimate}} & \multicolumn{3}{c}{CogVideoX \cite{yang2024cogvideox}} \\ \midrule
        Datasets & FVD ($\downarrow$) & FID ($\downarrow$) & CLIP ($\uparrow$) & FVD ($\downarrow$) & FID ($\downarrow$) & CLIP ($\uparrow$) & FVD ($\downarrow$) & FID ($\downarrow$) & CLIP ($\uparrow$) & FVD ($\downarrow$) & FID ($\downarrow$) & CLIP ($\uparrow$) & FVD ($\downarrow$) & FID ($\downarrow$) & CLIP ($\uparrow$) \\ \midrule
        HDTF & 177.88  & 28.89 &  {0.8912}
        & 175.01 & 25.89 & 0.9319 
        & 143.52 & 20.01 & 0.9128
        & 137.10  & 19.05 & 0.9236 
                        & 127.88 & 17.83 & 0.9247 \\ 
        TalkingHead-1KH & 261.16  & 32.67 &  0.8825 & 235.06 & 28.44 & 0.9133 & 176.43  & 21.56 & 0.9171 
        & 166.18  & 19.77 &  0.9212 & 167.23  &  19.02 & 0.9312 \\ 
        CelebV-HQ & 187.52  & 28.35 & 0.8925 & 169.43  & 25.05 & 0.9219 & 142.32 & 18.08 & 0.9374 & 147.62  & 17.45 & 0.9205 & 137.62  & 17.31 & 0.9305 \\ 
        CelebV-Text & 174.59  & 28.04 & 0.8874 & 159.52 & 26.83 & 0.9372 & 131.06 & 17.44 & 0.9355 & 121.22  & 16.53 & \textbf{0.9274} & 129.59  & 15.82 & 0.9388 \\ \midrule
        \textbf{DH-FaceVid-1K} & \textbf{156.13} & \textbf{26.69} & \textbf{0.9057} & \textbf{140.01}  & \textbf{24.95} & \textbf{0.9484} & \textbf{117.88} & \textbf{15.89}  & \textbf{0.9365} & \textbf{113.27}  & \textbf{13.91} & 0.9240 & \textbf{98.01}  & \textbf{11.73} & \textbf{0.9401} \\ \bottomrule
    \end{tabular}
    }
\vspace{-12pt}
\end{table*}


\begin{table*}[!htbp]
\LARGE
\setlength\tabcolsep{4pt}
    \centering
    \caption{Quantitative Results on Image-to-Video Generation. We evaluate the performance of five state-of-the-art open-source models for this task, ensuring consistent settings to facilitate fair comparison across different datasets.}
    \vspace{-8pt}
    \label{tab:quantitative-i2v}
    \resizebox{\textwidth}{!}{ 
    \begin{tabular}{l|ccc|ccc|ccc|ccc|ccc}
        \toprule
        \multicolumn{1}{l|}{Method} & \multicolumn{3}{c|}{SVD \cite{blattmann2023stable}} & \multicolumn{3}{c|}{I2VGen-XL \cite{2023i2vgenxl}} & \multicolumn{3}{c|}{ConsistI2V \cite{ren2024consisti2v}} & \multicolumn{3}{c|}{EasyAnimate \cite{xu2024easyanimate}} & \multicolumn{3}{c}{CogVideoX \cite{yang2024cogvideox}} \\ \midrule
        Datasets & FVD ($\downarrow$) & FID ($\downarrow$) & CLIP ($\uparrow$) & FVD ($\downarrow$) & FID ($\downarrow$) & CLIP ($\uparrow$) & FVD ($\downarrow$) & FID ($\downarrow$) & CLIP ($\uparrow$) & FVD ($\downarrow$) & FID ($\downarrow$) & CLIP ($\uparrow$) & FVD ($\downarrow$) & FID ($\downarrow$) & CLIP ($\uparrow$) \\ \midrule
        HDTF & 207.88  & 26.83  & - 
        & 240.19  & 28.05  & 0.9113 
        & 217.21  & 23.56  & 0.9057 
        & 135.06 & 21.71 & 0.9219 
        & 113.52 &15.14    & 0.9328 \\ 
        TalkingHead-1KH & 236.18  & 28.73 &  - 
        & 257.29  & 31.26 &  0.8962
        & 256.03  & 30.23 &  0.8989 
        & 151.01 & 20.01 & 0.9193 
        & 157.22  & 17.83   & 0.9399 \\ 
       CelebV-HQ & 187.99  & 25.70  & -
       & 217.62  & 26.60  & 0.9031
       & 225.79  & 24.95  & 0.9088 
       & 131.52  & 18.16 & 0.9335 
       & 130.91 & 16.25 & 0.9415 \\ 
        CelebV-Text & 162.59  & 23.07  & -
        & 185.59  & 23.07  & \textbf{0.9067 }
        & 172.89  & 23.21  &   0.8909
        & 125.14 & 17.54  & \textbf{0.9372} 
        & 123.25 & 14.83  & 0.9471 \\ \midrule
        \textbf{DH-FaceVid-1K} & \textbf{132.92}  & \textbf{18.95}  & -
        & \textbf{154.35}  & \textbf{18.61}  & 0.9013 &
        \textbf{152.92}  & \textbf{18.95}  & \textbf{0.9105} & \textbf{95.27}  & \textbf{13.71}  & 0.9321 & \textbf{92.31} & \textbf{10.69}  & \textbf{0.9514} \\ \bottomrule
    \end{tabular}
    }
\vspace{-12pt}
\end{table*}

\begin{table*}[!htbp]
\LARGE
    \centering
    \caption{
    Performance of the CogVideoX Model on the Text-to-Video and Image-to-Video Task. An investigation into the most cost-effective training data scale using varying model parameters and different data scales.}
    \label{tab:scale}
    \vspace{-8pt}
    \resizebox{\textwidth}{!}{ 
    \begin{tabular}{l|ccc|ccc|ccc|ccc}
        \toprule
        \multicolumn{1}{l|}{Task} & \multicolumn{6}{c|}{Text-to-Video} & \multicolumn{6}{c}{Image-to-Video} \\ \midrule
        \multicolumn{1}{l|}{Param.} & \multicolumn{3}{c|}{2B} & \multicolumn{3}{c|}{5B}  & \multicolumn{3}{c|}{2B} & \multicolumn{3}{c}{5B}\\ \midrule
        Data Scale & FVD ($\downarrow$) & FID ($\downarrow$) & CLIP ($\uparrow$) & FVD ($\downarrow$) & FID ($\downarrow$) & CLIP ($\uparrow$) 
        & FVD ($\downarrow$) & FID ($\downarrow$) & CLIP ($\uparrow$)
        & FVD ($\downarrow$) & FID ($\downarrow$) & CLIP ($\uparrow$)\\ \midrule
        100h & 215.06 & 17.01 & 0.9021 & 237.52 & 18.17 & 0.9128  
        & 159.26 & 14.01 & 0.9119
        & 183.52 & 14.17 & 0.9228\\
        200h & 185.06 & 16.23 & 0.9034 & 203.52 & 16.37 & 0.9156
        & 146.33 & 15.18 & 0.9232 & 162.44 & 15.73 & 0.9271\\ 
        400h & 177.18 & 14.16 & 0.9092 & 180.99 & 14.25 & 0.9197 
        & 147.18 & 13.06 & 0.9295 & 145.99 & 12.95 & 0.9314\\ 
        600h & \textbf{145.14} & \textbf{ 12.50} & \textbf{0.9173} & 143.25 & 12.83 & 0.9235 
       &\textbf{135.14} & \textbf{12.67} & \textbf{0.9306} & 133.85 & 12.83 & 0.9344\\ 
        800h & 148.82 & 13.15 & 0.9055 
        & 121.63 & 12.05 & 0.9385
        & 137.52 & 12.71 & 0.9242 & 113.63 & 11.90 & 0.9387\\ 
        1000h & 150.27 & 13.31 & 0.9108 & \textbf{98.01} & \textbf{11.73} & \textbf{0.9401}  
        & 141.36 & 12.94 & 0.9201 
        & \textbf{92.31} & \textbf{10.69} & \textbf{0.9514}\\ \bottomrule
    \end{tabular}
    }
\vspace{-12pt}
\end{table*}

These include thorough vetting of dataset usage requests, clear licensing terms, stringent rules governing the entire data lifecycle (from application to final disposal), and the removal of any personally identifiable information. 
To be specific, all users must submit an application form detailing research interests and purposes; following rigorous vetting, approved applicants enter a formal agreement that: (1) limits usage strictly to academic research, (2) prohibits dataset redistribution, (3) mandates transparent disclosure of research objectives and methodologies, and (4) requires secure data destruction upon project completion.

By championing responsible and open data usage, we aim to protect individual rights, respect societal norms, and foster a culture of accountability within the research community.


\begin{figure*}[t]
    \centering
    \includegraphics[width=\linewidth]{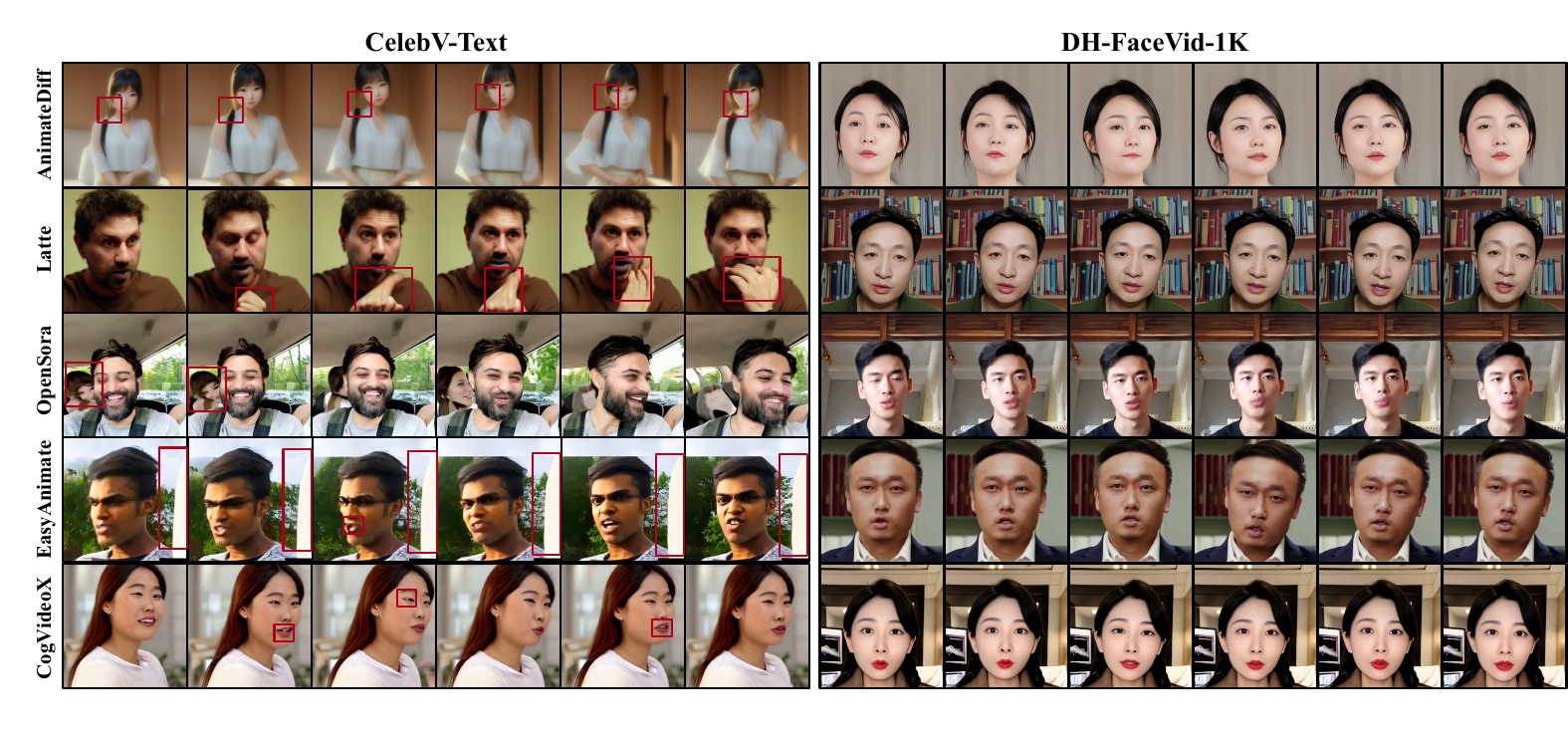}
    \vspace{-18pt}
    \caption{\textbf{Text-to-Video.} 
    Using the same settings across different methods, we compare models trained on CelebV-Text and our collected dataset (DH-FaceVid-1K). For qualitative comparison, we use the exact same text prompts (\ie, \textit{``A young man is talking.''} and \textit{``A beautiful young woman is talking.''}). Please zoom in to observe the facial details closely.}
    \vspace{-7pt}
    \label{fig:t2v}
 \end{figure*}

\begin{figure*}[t]
    \centering
    \includegraphics[width=\linewidth]{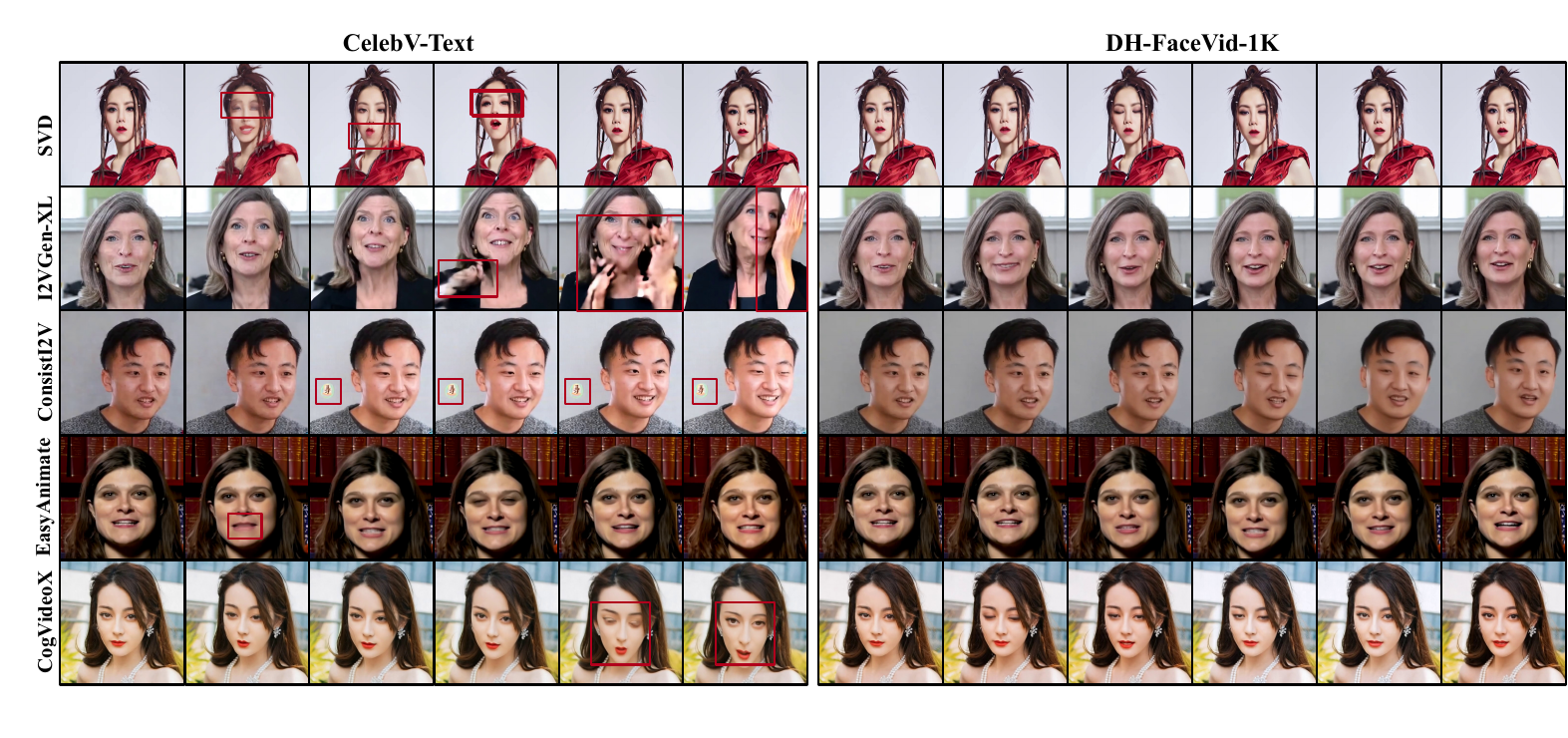}
    \vspace{-18pt}
    \caption{\textbf{Image-to-Video.}
    Using the same settings across different methods, we compare models trained on CelebV-Text and our collected dataset (DH-FaceVid-1K). For qualitative comparison, we use the exact same text prompts (\ie, \textit{``A young man is talking.''} and \textit{``A beautiful young woman is laughing.''}) as well as the same image prompt (the initial image). 
    }
    \vspace{-7pt}
    \label{fig:i2v}
\end{figure*}


\section{Experiments}

In this section, we conducted experiments on public face video datasets and our collected DH-FaceVid-1K dataset, focusing on two popular video generation tasks: Text-to-Video and Image-to-Video generation.
We employed three widely accepted evaluation metrics, namely FVD \cite{unterthiner2018towards}, FID \cite{heusel2017gans}, and CLIPScore \cite{hessel2021clipscore} (abbreviated as ``CLIP'') to assess temporal consistency, single frame quality, and video-text similarity, respectively.
For the construction of the reference dataset used in the following experiments, we followed CelebV-Text's protocol by randomly sampling 2,048 videos from the non-training portions of the datasets listed in Table \ref{tab:meta}.
Both qualitative and quantitative results demonstrated the effectiveness and superiority of the proposed DH-FaceVid-1K in these human face video generation tasks.
Additionally, we draw empirical insights to answer the two research questions (\ie, RQ1: Analysis for Data Scale Requirement and RQ2: Difference of Backbones) raised in previous sections.

\vspace{-4pt}
\subsection{Experiments on Text-to-Video Generation}
We conducted experiments to demonstrate the effectiveness of DH-FaceVid-1K for Text-to-Video generation using state-of-the-art open-source methods, namely AnimateDiff \cite{guo2023animatediff}, Latte \cite{ma2024latte}, OpenSora \cite{opensora}, EasyAnimate \cite{xu2024easyanimate}, and CogVideoX \cite{yang2024cogvideox}.

We fine-tuned these models using the datasets specified in Table \ref{tab:quantitative-t2v} and the official implementation code.
The hyperparameters varied slightly across model configurations. 
Settings like batch size and learning rate were based on optimal parameters from official implementations and technical reports.
Specifically, for CogVideoX, the parameters included a batch size of 1, a learning rate of 1e-5, and 30k steps.
As shown in Figure \ref{fig:t2v}, we benchmarked our dataset against the highest-quality public dataset, CelebV-Text, using the same settings and text prompts.

Models trained on CelebV-Text struggle with Asian face quality and produce artifacts like ``random hand noise" and ``multiple faces".
In contrast, models trained on DH-FaceVid-1K do not exhibit these issues and show superior detail in areas like lips and teeth.
As shown in Table \ref{tab:quantitative-t2v}, while all models trained on other datasets perform competitively across metrics, those trained on our dataset significantly outperform the others.
Specifically, the best-performing model in our experiments, CogVideoX \cite{yang2024cogvideox}, trained on our dataset, exhibits notably enhanced performance.

This superior performance can be attributed to the rigorous filtering of low-quality data, which enhances the training outcomes.

\vspace{-4pt}

\subsection{Experiments on Image-to-Video Generation}
We used the open-source, widely-adopted Image-to-Video models Stable Video Diffusion (SVD) \cite{blattmann2023stable}, I2VGen-XL \cite{2023i2vgenxl}, ConsistI2V \cite{ren2024consisti2v}, EasyAnimate \cite{xu2024easyanimate}, and CogVideoX \cite{yang2024cogvideox} to compare the performance of models trained on public datasets with those trained on our dataset.
We implemented these methods using the datasets specified in Table \ref{tab:quantitative-t2v} and the official implementation code.
The hyperparameters varied slightly for different model configurations, based on their official implementations and technical reports.
Specifically, for CogVideoX, the parameters included a batch size of 1, a learning rate of 1e-5, and 30k steps.
The training configurations for other models also followed their official repos.
The qualitative comparison results, illustrated in Figure \ref{fig:i2v}, show that models trained on our dataset outperform those trained on the leading public dataset.
Since images inherently contain more detailed information compared to ambiguous prompts of natural language descriptions, all models can generate Asian face videos that retain the input face characteristics.
However, the models trained with DH-FaceVid-1K still demonstrate superior performance in capturing facial details such as teeth and eyes.
They also generate a wider range of natural facial motions and expressions due to the higher quality and diversity of our dataset.
The quantitative results presented in Table \ref{tab:quantitative-i2v} demonstrate that all these models show competitive scores across three metrics when trained on different datasets.


\subsection{RQ1: Analysis for Data Scale Requirement}
We conducted a series of experiments to investigate this research question, using CogVideoX \cite{yang2024cogvideox} for both T2V and I2V tasks.
We randomly sampled a series of videos (\ie, 100h, 200h, 400h, 600h, 800h, and 1,000h) from our collected dataset and performed identical training setups with different parameter configurations of the model.
All models were trained with a learning rate of 1e-5, a batch size of 1, and other identical hyperparameters for 30k steps.
For the test set, we separately sampled 2,048 videos before dividing the complete dataset into subsets of different durations. 
From each sampled video, we uniformly extracted 4 frames.
2,048 videos and 8,912 frames were used as the GT for calculating FVD and FID, respectively.

As shown in Table \ref{tab:scale}, smaller models perform well on limited data but struggle with larger, more diverse datasets.
This is reflected in the lack of growth or slight decline of metrics.
Specifically, we observe that a 2B-parameter model trained on a 100-hour dataset produced highly similar videos when given similar prompts.
Through these experiments, we conclude that for a 2B-scale DiT-based backbone face video generation model (\ie, CogVideoX \cite{yang2024cogvideox}), a dataset size of approximately 600 hours strikes a good balance between diversity and quality in the generated videos for fine-tuning.
On the other hand, the metrics of larger models on small datasets are slightly lower than those of their smaller counterparts.
For instance, a 6B-parameter model fine-tuned on the optimal 600-hour dataset configuration for the 2B model fails to achieve comparable performance.
The training process was unstable, the generated results often contain noise, and the model's performance does not improve as steadily as expected as training progresses.
We attribute this to a mismatch between the dataset size and the parameter scale of the model.
By analyzing FVD, FID, and CLIPScore and manual evaluation, we reveal that the 6B-parameter model fine-tuned on the full DH-FaceVid-1K, or even a larger dataset, performs best.

\vspace{-4pt}

\subsection{RQ2: Analysis for Difference of Backbones}
For this research question, we primarily examine the quantitative and qualitative results from previous experiments to draw empirical conclusions about training domain-specific video generation models from scratch or fine-tuning pre-trained open-world video generation models.
From the results of both Text-to-Video and Image-to-Video tasks, we observe that Diffusion-Transformer (DiT) based models (\eg, Latte \cite{ma2024latte}, CogVideoX \cite{yang2024cogvideox}) usually outperform UNet-based models (\eg, AnimateDiff \cite{guo2023animatediff}).
The experimental results show that UNet-based models (\eg, SVD \cite{blattmann2023stable}) converge faster than DiT-based ones, while the latter require extensive training data, time, and GPU resources due to their higher computational demands.
Among them, CogVideoX excels with its Expert Transformer architecture and progressive training strategy, delivering higher-quality, more diverse outputs and superior fine-tuning performance.
Our conclusion aligns with the Hallo3 \cite{cui2025hallo3} (DiT) versus Hallo2 \cite{cui2024hallo2} (UNet) comparison: while the former achieves superior performance metrics (\eg, lower FVD), it demands significantly more training resources than the latter.


\vspace{-6pt}

\section{Conclusion}
In this work, we introduce a large-scale, high-quality face video dataset, DH-FaceVid-1K.
We hope that this resource can meet the demands of research tasks related to human-centric video generation.
We conducted extensive experiments on the dataset, yielding several valuable empirical insights.
In future work, we will expand the dataset to improve demographic diversity (\eg, individuals from Indonesia and Vietnam) and release additional pre-trained video generation models to benefit the community.
\section{Acknowledgements}
This work was supported by the National Natural Science Foundation of China (NSFC) grant U22A2094.

{
    \small
    \bibliographystyle{ieeenat_fullname}
    \bibliography{main}
}

\clearpage

\begin{table*}
\centering
\caption{Comprehensive attribute list of DH-FaceVid-1K, including ethnicities, appearance details, emotions, actions, and lighting conditions.}
\label{tab:attr_list}
\resizebox{0.72\textwidth}{!}{%
\begin{tabular}{llllll} 
\toprule
\multicolumn{6}{c}{\textbf{Static Attributes}} \\ 
\hline
\multicolumn{6}{c}{\textbf{(a) Ethnicities}} \\ 
Asian & White & Indian & European & African & Arab \\ 
Latino & Chinese & Japanese & Korean & Latina & Jewish \\ 
Mexican & & & & & \\ \hline

\multicolumn{6}{c}{\textbf{(b) General Appearance}} \\ 
No beard & Male & Female & Straight hair & Black hair & Anchored eyebrow \\ 
Short hair & Eyeglasses & Brown hair & Rosy cheeks & Long hair & Bush eyebrow \\ 
Chubby & Heavy makeup & Oval face & Lipsticks & Big nose & Wavy hair \\ 
Bangs & Double chin & Side burn & Earrings & Goatee & White hair \\ 
Blonde hair & Bald & Old & Narrow eyes & Big lips & Pale skin \\ 
Wearing shirt & Wearing jacket & Wearing suit & Wearing tie & Wearing coat & Wearing blazer \\ 
Wearing hoodie & Redhead & Wearing cloak & Wearing bonnet & Blemish & Pompadour \\ 
Mustache & Wearing necklace & Wearing sweater & Wearing blouse & Wearing cap & Wearing scarf \\ 
Wearing hat & Wearing dress & Wearing headband & Mole & Wearing vest & Tattoos \\ 
Wearing sunglasses & Piercings & Wearing beanie & Wearing hijab & Ponytail & Scars \\ 
Wearing robe & Tan & Wearing ring & Wrinkles & Wearing hood & Wearing backpack \\ 
Wearing bandana & Braids & Wearing turban & Pigtails & Dreadlocks & With purse \\ \hline

\multicolumn{6}{c}{\textbf{(c) Light Conditions}} \\ 
Dark & Outdoor & Bright & Natural & Artificial & Dim \\ 
Daylight & Normal & & & & \\ \midrule

\multicolumn{6}{c}{\textbf{Dynamic Attributes}} \\ 
\hline
\multicolumn{6}{c}{\textbf{(a) Action}} \\ 
Talk & Smile & Wag Head & Frown & Gaze & Nod \\ 
Shake Head & Sing & Glare & Close Eyes & Laugh & Whisper \\ 
Wink & Shout & Sigh & Listen To Music & Yawn & Cough \\ 
Sneeze & Study & Speak & Sleep & Kiss & Pile \\ 
Stand & Think & Eat & Drink & Read & Touch \\ 
Roll & Rest & Work & Wash & Ride & Drive \\ 
Play & Hold & Throw & Run & Clutch & Tilt \\ 
Hug & Walk & & & & \\ \hline

\multicolumn{6}{c}{\textbf{(b) Emotion}} \\ 
Neutral & Happy & Sad & Angry & Fear & Surprise \\ 
Contempt & Disgust & & & & \\ 
\bottomrule
\end{tabular}}
\end{table*}

\section*{Dataset Construction}
\subsection*{A. Data details.}
The majority of the facial video data in this dataset, over 65\%, consists of single-person frontal interview videos, ensuring the quality and stability of the raw data. To enhance the diversity of video scenes, we also collect some outdoor vlog-style talking head video data. To ensure the diversity of the dataset, in over 95\% of total data, each identity corresponds to no more than 10 video clips.

As mentioned in the manuscript, we applied the data processing pipeline from the text to filter the public datasets CelebV-HQ, CelebV-Text, and TalkingHead-1KH. The filtered sample data can be found in Figure \ref{fig:filtered}.

\subsection*{B. CodeFormer bias}
\begin{figure}[htbp]
    \centering
    \includegraphics[width=\linewidth]{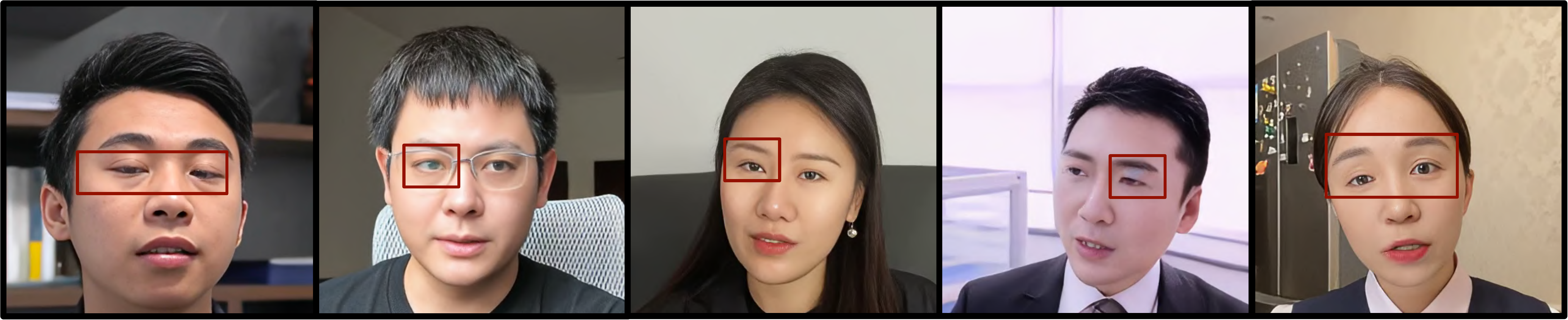}
    \caption{Sample image showing artifacts in the eye area after introducing CodeFormer.}
    \label{fig:codeformers}
\end{figure}

\begin{figure}[htbp]
    \centering
    \includegraphics[width=\linewidth]{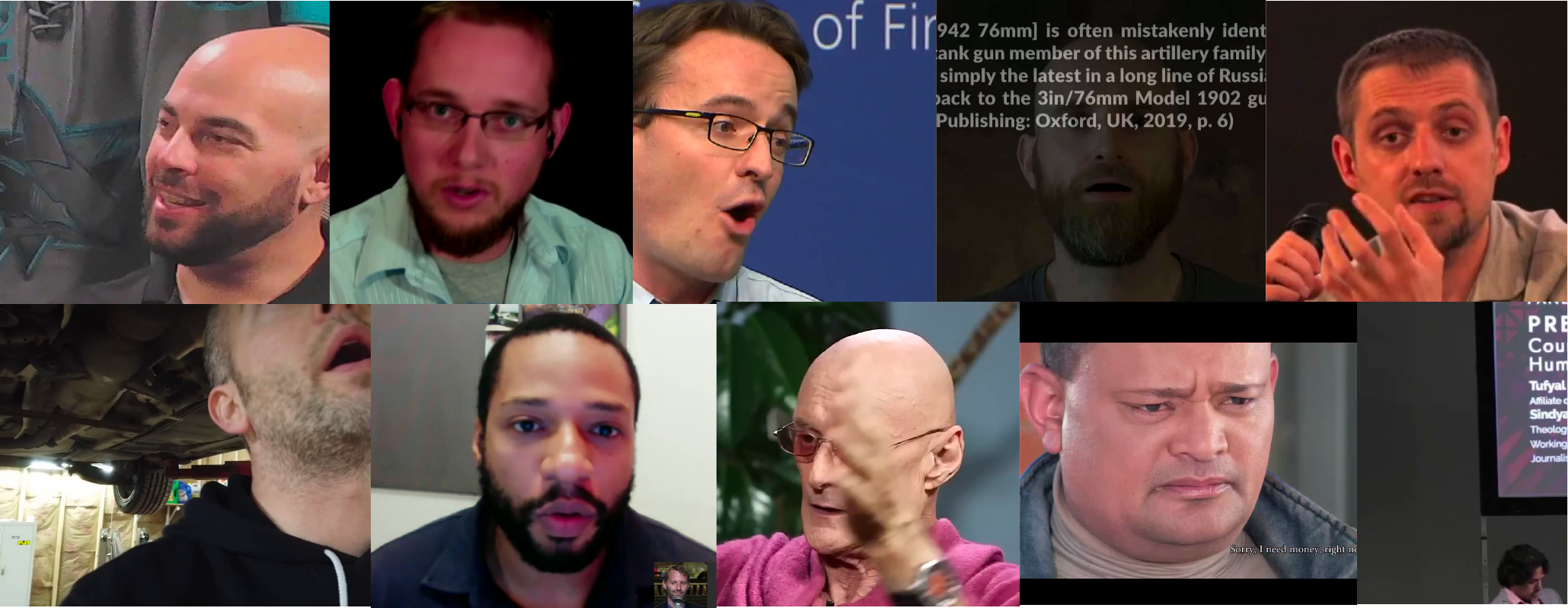}
    \caption{Filtered out sample data from public dataset.}
    \label{fig:filtered}
\end{figure}

CodeFormer-enhanced videos sometimes exhibit artifacts, particularly around the eyes of Asian faces. To address this, a team of 10 manually reviews the first frame of each video, discarding any that display such issues. The demo data corresponding to these failures can be found in Figure \ref{fig:codeformers}.

\begin{figure}[htbp]
    \centering
    \includegraphics[width=\linewidth]{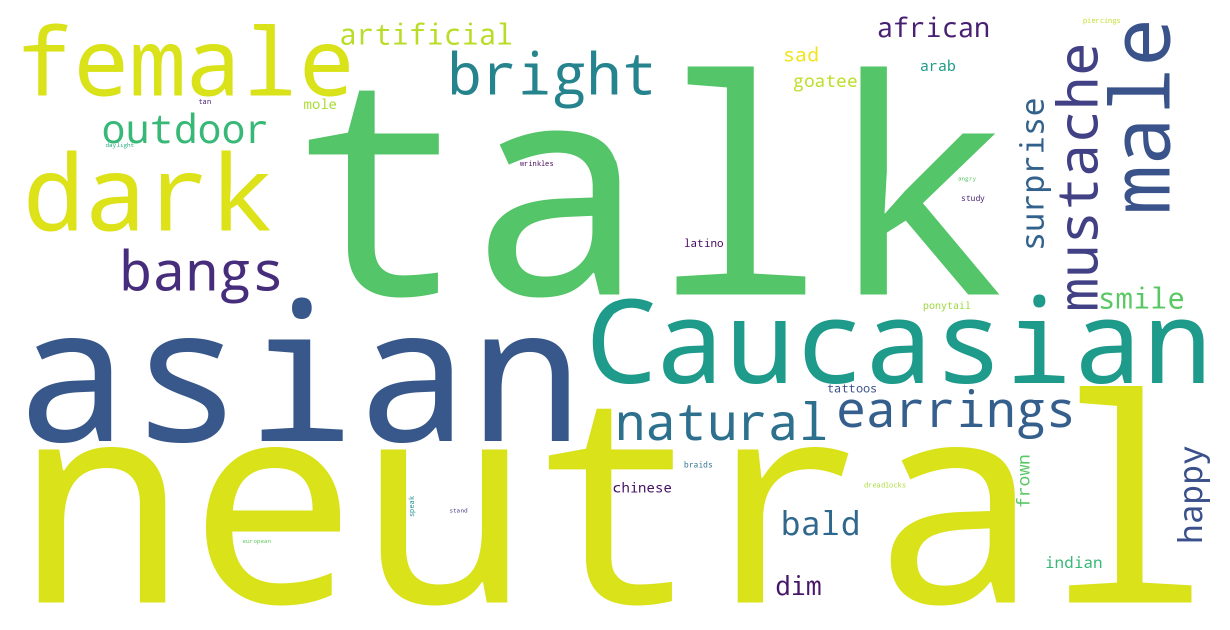}
    \caption{Key words of DH-FaceVid-1K's annotations.}
    \label{fig:wordcloud}
\end{figure}
\begin{figure}[htbp]
    \centering
    \includegraphics[width=\linewidth]{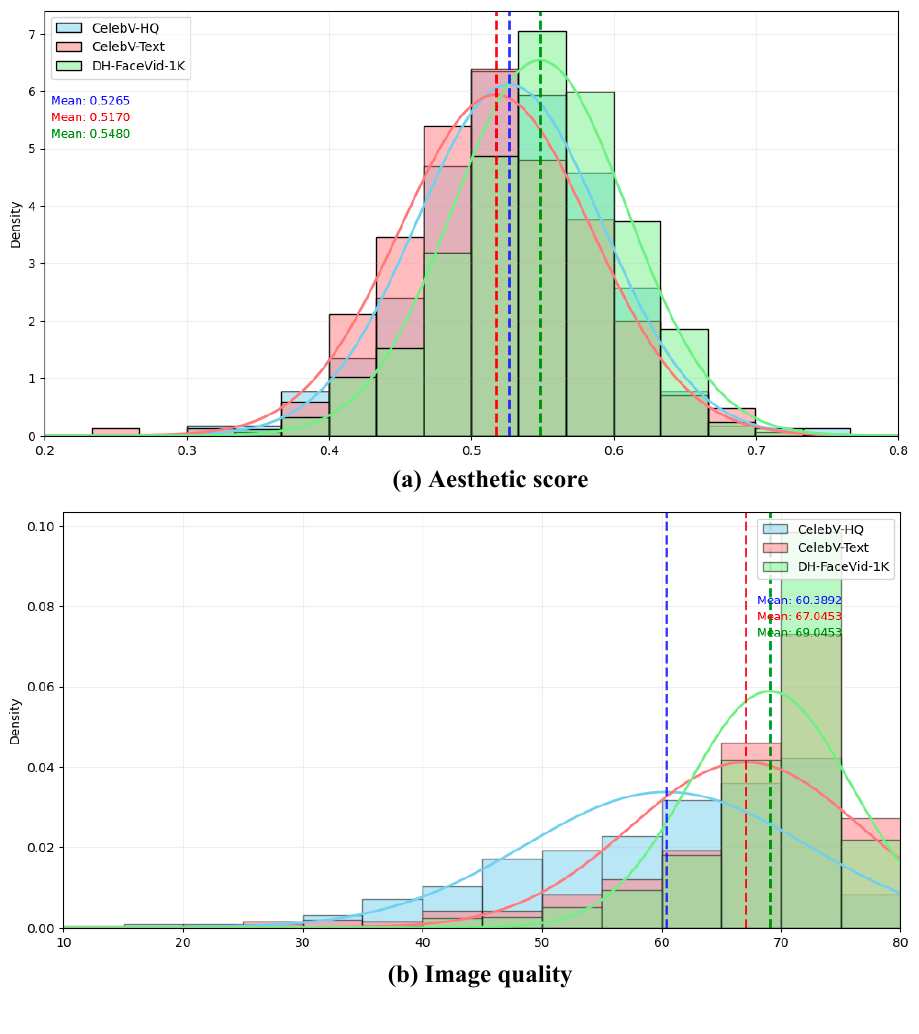}
    \caption{Comparison of CelebV-HQ, CelebV-Text and \textbf{DH-FaceVid-1K} on VBench.}
    \label{fig:compare}
\end{figure}
\subsection*{C. Video Annotations}
We first use PLLaVA to automatically generate video annotations. More specifically, we interact with PLLaVA in a manner similar to conversing with an LLM (Large Language Model). We provide it with a template prompt, which it then uses to generate corresponding video annotations. Our annotations primarily include information about appearance, expressions, actions, and the environment—details that are also embedded within the template, where we provide options for various attributes of these annotations. For example, for the attribute ``emotion'',  we require  selecting from options such as Neutral, Happy, Sad, and Angry. The selection of these options is based on the attribute design from CelebV-Text.

After obtaining the initial annotations, we employ an additional 50 annotators to review the text. These annotators are tasked with assessing static information in the annotations, such as age, ethnicity, clothing, and facial accessories, based on the first frame of the video. Subsequently, they evaluate dynamic information, such as emotional and movement changes, based on three randomly selected non-consecutive additional frames.

The comprehensive attribute list of DH-FaceVid-1K is presented in Table \ref{tab:attr_list}
We visualized some high-frequency words in the annotations. The word cloud can be found in Figure \ref{fig:wordcloud}
\subsection*{D. Challenges}
Although our primary goal is to construct a high-quality talking face dataset mainly composed of Asians to address the scarcity of such datasets, the statistical data presented in the main text indicates a certain long-tail distribution in racial representation. However, through experiments, we discover that by integrating our proposed dataset with cleaned open-source data and thoroughly training powerful video generation models, such as CogVideoX, it is still easy to generate high-quality data for Caucasian, African, and Indian individuals.
In terms of actions and emotions, talking and neutral emotions dominate overwhelmingly. Nevertheless, we find that even a small proportion of data is sufficient for the model to learn, given our total duration of nearly 1200 hours. Qualitative experimental images for other ethnicities, scenes, emotions, and actions, such as smiling, can be found in the later sections.

\section*{E. Training Protocol }
To validate the effectiveness and superiority of proposed dataset, we evaluated the performance of generative models under different architectures based on T2V (Text-to-Video) and T2I (Text-to-Image) in the earlier experimental sections.
All model training is conducted separately using CelebV-Text and DH-FaceVid-1K as the training datasets.

We strive to replicate these methods according to the training details provided officially, including the optimizer, learning rate, and loading the official weight checkpoints. All models are fine-tuned for 100k steps using eight A800 GPUs. However, for SVD, since the official training code is not open-sourced, we train it based on a third-party reproduced code.

For OpenSora, we fine-tune using version 1.2. For EasyAnimate, we use version v3 for fine-tuning. However, this version often encounters crashes when fine-tuning with large datasets. Our solution is to reduce the learning rate (down to 2e-6) and frequently save checkpoints, allowing us to revert to the most recent stable checkpoint in case of a crash.
\section*{F. Quantative Experiment Settings}
We utilize FVD, FID, and CLIPSIM to evaluate video temporal consistency, individual frame quality, and the relevance between the generated video and text prompt. Before training different models, we randomly select 2048 videos from each of the two training datasets as the test set, which are not involved in training. For Text-to-Video (T2V) generation experiments, we generate 2048 videos as the ``fake'' videos to calculate FVD and CLIPSIM, based on the prompts of the test set. Subsequently, to compute FID, we uniformly sample 4 frames from each of these 2048 ``fake'' videos as fake data and apply the same procedure to the test data. In total, we have 8192 images for both real and fake data. For Image-to-Video (I2V), we take the first frame from the test set images as the image prompt to generate 2048 videos as fake data, with other settings similar to the T2V experiments.

To ensure the fairness of the comparison, all real and fake data are uniformly resized to 512 $\times$ 512.

\section*{G. More Qualitative Results}
We present additional qualitative results that were not covered in the main text. For the Text-to-Video experiments, we first use more diverse prompts, such as different emotions, ethnicities, actions, and scenes, to generate results. We then include generation results from prompts with formats different from those in the dataset annotations, such as prompts lacking key information or using single-word descriptions of character traits in the video. This demonstrates that our model, after thorough training, possesses strong generalization capabilities. For Image-to-Video (I2V), we use some``wild'' data to generate results, verifying that the model, trained on our dataset, still exhibits good extrapolation abilities. The additional qualitative experimental results in this section are generated using CogVideoX 5b version and Open Sora.

\section*{H. Comparison using Vbench}
We conducted a comparative analysis of aesthetic scores and image quality metrics between DH-FaceVid-1K, CelebV-HQ, and CelebV-Text using VBench \cite{huang2023vbench}, as illustrated in Figure \ref{fig:compare}. The results demonstrate that the proposed DH-FaceVid-1K dataset achieves comparable aesthetic performance while exhibiting superior image quality relative to the benchmark datasets.

\section*{I. Combination with Existing Datasets}
Our dataset includes 200h of Caucasian and African data, with diverse generation examples shown in Appendix Figures 7-8.
Given that existing public datasets already contain extensive non-Asian data (about 4.5k h), we recommend using either equal-duration sampling (1:1) or \textbf{population-proportional splits} to ensure generalization and racial diversity.

\begin{table}[t]
\centering
\caption{Comparison of T2V metrics on cross-dataset.}
\vspace{-1em}
\label{tab:re-quantitative-t2v}
\resizebox{0.88\columnwidth}{!}{%
\begin{tabular}{@{}lcccc@{}}
\toprule
Dataset & \multicolumn{1}{c}{FID ($\downarrow$)} & \multicolumn{1}{c}{FVD ($\downarrow$)} & \multicolumn{1}{c}{CLIP ($\uparrow$)} & \multicolumn{1}{c}{DSL-FIQA ($\uparrow$)} \\ 
\midrule
CelebV-Text & 25.71 & 313.51 & 0.838 & 0.7531 \\ 
Ours & \textbf{22.81} & \textbf{250.53} & \textbf{0.845} & \textbf{0.8204} \\ 
\bottomrule
\end{tabular}
}
\vspace{-1.5em}
\end{table}

\section*{J. Cross Reference-Dataset Evaluation}
We have now conducted cross-dataset evaluation using CogVideoX-2B T2V to assess the generalization capability of our dataset. The evaluation sets include HDTF and CelebV-HQ (1:3, results are reported in Table \ref{tab:re-quantitative-t2v} here).
Our dataset demonstrates strong generalizability and consistently outperforms CelebV-Text across all metrics. 

\section*{K. User Study}
We conduct a user study with 25 participants to evaluate   CogVideoX's T2V face generation performance after fine-tuning. Participants compare 30 videos generated with the same prompt by models trained on CelebV-HQ, CelebV-Text, and the proposed dataset. The prompts include diverse ethnic groups and emotions(neutral:happy:angry:sad = 6:1:1:1) in various settings. Evaluations focus on prompt adherence, face naturalness, video quality, and overall quality. As shown in Table \ref{tab:user-study}, videos using our fine-tuned weights score the highest overall.

\begin{table}[t]
    \small
    \setlength{\tabcolsep}{3pt}  
    \centering 
    \caption{User study measured by Mean Opinion Scores.}
    \vspace{-0.5em}  
    \begin{tabular}{lcccc} 
        \toprule
        Datasets  & Adherence & Naturalness  & Visual Quality & Overall\\ 
        \midrule
        CelebV-HQ  & 2.53  & 4.01 & 3.75   &3.90\\
        CelebV-Text  & 2.89  & 3.88 & 4.17   &4.01\\
        Ours & \textbf{4.52}  & \textbf{4.39}  &\textbf{4.33}   &\textbf{4.47}\\
        \bottomrule
    \end{tabular}
    \vspace{-2em}  
    \label{tab:user-study}
\end{table}

\section*{L. Future Work}
Our proposed dataset primarily focuses on talking faces, and when combined with powerful open-source models, it already enables robust and diverse talking face generation capabilities. However, there are still limitations, such as the video's restriction to facial expressions without including common actions like eating or crying. 
On the other hand, our current dataset contains a limited number of non-East Asian faces. To further enhance data diversity, we are actively collecting additional Asian facial data from regions beyond East Asia (including Indian, Indonesian, Malay, and Vietnamese populations), while strictly adhering to data privacy protocols.

Our future work will focus on developing datasets that support digital human generation for both half-body and full-body models, aiming to achieve true digital human synthesis.
\begin{figure*}[t]
\centering
\includegraphics[width=\textwidth]{figures/supple-t2v-1.pdf}
\vspace{-10pt}
\caption{More qualitative T2V evaluation results from CogVideoX demonstrate the effectiveness of the DH-FaceVid-1K fine-tuned model in generating capabilities from different angles and facial distances}
\vspace{-15pt}
\label{fig:t2v-1}
\end{figure*}
\clearpage

\begin{figure*}[t]
\centering
\includegraphics[width=\textwidth]{figures/supple-t2v-2.pdf}
\vspace{-10pt}
\caption{More qualitative T2V evaluation results from OpenSora confirm that the DH-FaceVid-1K fine-tuned model retains the ability to generate speaking videos of faces from multiple ethnicities, including Caucasian, African, and Indian.}
\vspace{-15pt}
\label{fig:t2v-2}
\end{figure*}
\clearpage

\begin{figure*}[t]
\centering
\includegraphics[width=\textwidth]{figures/supple-i2v-1.pdf}
\vspace{-10pt}
\caption{More qualitative I2V evaluation results from CogVideoX confirm that the DH-FaceVid-1K fine-tuned model retains the ability to generate various styles (realistic and cartoon) across multiple ethnicities.}
\vspace{-15pt}
\label{fig:i2v-1}
\end{figure*}
\clearpage

\begin{figure*}[t]
\centering
\includegraphics[width=\textwidth]{figures/supple-i2v-2.pdf}
\vspace{-10pt}
\caption{More qualitative I2V evaluation results from CogVideoX confirm that the DH-FaceVid-1K fine-tuned model retains the ability to generate not only speaking but also various other actions, such as reading, and can produce half-body images beyond just the face, across multiple ethnicities.}
\vspace{-15pt}
\label{fig:i2v-2}
\end{figure*}

\end{document}